\documentclass[11pt,a4paper]{article}
\usepackage[hyperref]{emnlp-ijcnlp-2019}
\usepackage{times}
\usepackage{latexsym}
\usepackage{graphicx}
\usepackage{url}
\usepackage{enumitem}
\usepackage{amsfonts}
\usepackage{amsmath}

\usepackage{xcolor}

\graphicspath{ {.} }

\setenumerate{noitemsep,topsep=10pt,parsep=0pt,partopsep=0pt}
\aclfinalcopy 

\DeclareMathOperator{\minMax}{minMax}

\title{Exploring Multilingual Syntactic Sentence Representations}

\author{Chen Liu, Anderson de Andrade, Muhammad Osama \\
	Wattpad \\
	Toronto, ON, Canada \\
  \texttt{cecilia, anderson, muhammad.osama@wattpad.com}
}

\date{}

\begin{document}
\maketitle
\begin{abstract}
We study methods for learning sentence embeddings with syntactic structure. We focus on methods of learning syntactic sentence-embeddings by using a multilingual parallel-corpus augmented by Universal Parts-of-Speech tags. We evaluate the quality of the learned embeddings by examining sentence-level nearest neighbours and functional dissimilarity in the embedding space. We also evaluate the ability of the method to learn syntactic sentence-embeddings for low-resource languages and demonstrate strong evidence for transfer learning. Our results show that syntactic sentence-embeddings can be learned while using less training data, fewer model parameters, and resulting in better evaluation metrics than state-of-the-art language models.
\end{abstract}

\section{Introduction}

Recent success in language modelling and representation learning have largely focused on learning the semantic structures of language \cite{BERT}. Syntactic information, such as part-of-speech (POS) sequences, is an essential part of language and can be important for tasks such as authorship identification, writing-style analysis, translation, etc. Methods that learn syntactic representations have received relatively less attention, with focus mostly on evaluating the semantic information contained in representations produced by language models.

Multilingual embeddings have been shown to achieve top performance in many downstream tasks \cite{conneau2017word, artetxe2018}. By training over large corpora, these models have shown to generalize to similar but unseen contexts. However, words contain multiple types of information: semantic, syntactic, and morphologic. Therefore, it is possible that syntactically different passages have similar embeddings due to their semantic properties. On tasks like the ones mentioned above, discriminating using patterns that include semantic information may result in poor generalization, specially when datasets are not sufficiently representative. 

In this work, we study methods that learn sentence-level embeddings that explicitly capture syntactic information. We focus on variations of sequence-to-sequence models \cite{seq2seq}, trained using a multilingual corpus with universal part-of-speech (UPOS) tags for the target languages only. By using target-language UPOS tags in the training process, we are able to learn sentence-level embeddings for source languages that lack UPOS tagging data. This property can be leveraged to learn syntactic embeddings for low-resource languages.

Our main contributions are: to study whether sentence-level syntactic embeddings can be learned efficiently, to evaluate the structure of the learned embedding space, and to explore the potential of learning syntactic embeddings for low-resource languages.

We evaluate the syntactic structure of sentence-level embeddings by performing nearest-neighbour (NN) search in the embedding space. We show that these embeddings exhibit properties that correlate with similarities between UPOS sequences of the original sentences. We also evaluate the embeddings produced by language models such as BERT \cite{BERT} and show that they contain some syntactic information.

We further explore our method in the few-shot setting for low-resource source languages without large, high quality treebank datasets. We show its transfer-learning capabilities on artificial and real low-resource languages.

Lastly, we show that training on multilingual parallel corpora significantly improves the learned syntactic embeddings. This is similar to existing results for models trained (or pre-trained) on multiple languages \cite{minning, artetxe2018} for downstream tasks \cite{XlangPretrain}.

\section{Related Work}

Training semantic embeddings based on multilingual data was studied by MUSE \cite{conneau2017word} and LASER \cite{artetxe2018} at the word and sentence levels respectively. Multi-task training for disentangling semantic and syntactic information was studied in \cite{chen2019}. This work also used a nearest neighbour method to evaluate the syntactic properties of models, though their focus was on disentanglement rather than embedding quality.


The syntactic content of language models was studied by examining syntax trees \cite{hewitt2019}, subject-object agreement \cite{yoav2019}, and evaluation on syntactically altered datasets \cite{linzen2016, marvin2018}. These works did not examine multilingual models.

Distant supervision \cite{meng2016, plank2018} has been used to learn POS taggers for low-resource languages using cross-lingual corpora. The goal of these works is to learn word-level POS tags, rather than sentence-level syntactic embeddings. Furthermore, our method does not require explicit POS sequences for the low-resource language, which results in a simpler training process than distant supervision.

\section{Method}\label{method}


\subsection{Architecture}

We iterated upon the model architecture proposed in LASER \cite{artetxe2018}. The model consists of a two-layer Bi-directional LSTM (BiLSTM) encoder and a single-layer LSTM decoder. The encoder is language agnostic as no language context is provided as input. In contrast to LASER, we use the concatenation of last hidden and cell states of the encoder to initialize the decoder through a linear projection.

\begin{figure*}
	\centering
	\includegraphics[width=\linewidth, trim={1cm 1cm 1cm 1cm}]{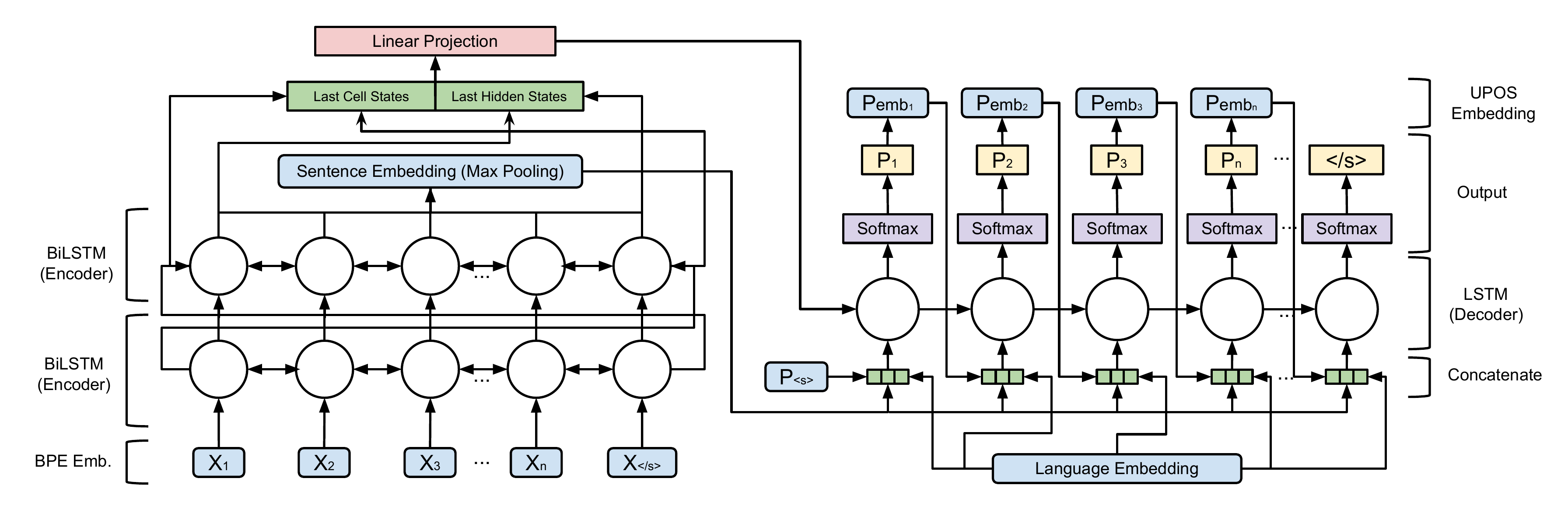}
	\caption{Proposed architecture.}
	\label{model}
\end{figure*}

At each time-step, the decoder takes an embedding of the previous POS target concatenated with an embedding representing the language context, as well as a max-pooling over encoder outputs. Figure \ref{model} shows the architecture of the proposed model.

The input embeddings for the encoder were created using a jointly learned Byte-Pair-Encoding (BPE) vocabulary \cite{sennrich-etal-2016-neural} for all languages by using sentencepiece\footnote{https://github.com/google/sentencepiece}. 

\begin{table}
	\caption{Hyperparameters}
	\label{param_table}
	\centering
	\begin{tabular}{ll}
		\hline
		Parameter & Value \\
		\hline
		Number of encoder layers & 2 \\
		Encoder forward cell size & 128 \\
		Encoder backward cell size & 128 \\
		\hline
		Number of decoder layers & 1 \\
		Decoder cell size & 512 \\
		\hline
		Input BPE vocab size & 40000 \\
		BPE embedding size & 100 \\
		UPOS embedding size & 100 \\
		Language embedding size & 20 \\
		Dropout rate & 0.2 \\
		Learning rate & 1e-4 \\
		Batch size & 32 \\
		\hline
	\end{tabular}
\end{table}

\subsection{Training}

Training was performed using an aligned parallel corpus. Given a source-target aligned sentence pair (as in machine translation), we:

\begin{enumerate}
	\item Convert the sentence in the source language into BPE
	\item Look up embeddings for BPE as the input to the encoder
	\item Convert the sentence in a target language into UPOS tags, in the tagset of the target language.
	\item Use the UPOS tags in step 3 as the targets for a cross-entropy loss.
\end{enumerate}

Hence, the task is to predict the UPOS sequence computed from the translated input sentence.

The UPOS targets were obtained using StandfordNLP \cite{qi2018universal} \footnote{https://stanfordnlp.github.io/stanfordnlp/index.html}. Dropout with a drop probability of 0.2 was applied to the encoder. The Adam optimizer \cite{ADAM} was used with a constant learning rate of $0.0001$. Table \ref{param_table} shows a full list of the hyperparameters used in the training procedure. 

\subsection{Dataset}
\label{subsec:dataset}



To create our training dataset, we followed an approach similar to LASER. The dataset contains 6 languages: English, Spanish, German, Dutch, Korean and Chinese Mandarin. These languages use 3 different scripts, 2 different language orderings, and belong to 4 language families. 

English, Spanish, German, and Dutch use a Latin-based script. However, Spanish is a Romantic language while the others are Germanic languages.  Chinese Mandarin and Korean are included because they use non-latin based scripts and originate from language families distinct from the other languages. Although the grammatical rules vary between the selected languages, they share a number of key characteristics such as the \textit{Subject-Verb-Object} ordering, except Korean (which mainly follows the \textit{Subject-Object-Verb} order). We hope to extend our work to other languages with different scripts and sentence structures, such as Arabic, Japanese, Hindi, etc. in the future.

The dataset was created by using translations provided by Tatoeba\footnote{https://tatoeba.org/eng/} and OpenSubtitles\footnote{http://opus.nlpl.eu/OpenSubtitles-v2018.php} \cite{opensrt}. They were chosen for their high availability in multiple languages. 

Statistics of the final training dataset are shown in Table~\ref{data_stats}. Rows and columns correspond to source and target languages respectively.

\begin{table*}[t!]
	\centering
	\caption{Training Dataset Statistics}
	\label{data_stats}
	\centering
	\begin{tabular}{lllllll}
		\hline
		{} & English & German & Spanish & Chinese & Korean & Dutch   \\
		\hline
		English & - & 521.87k&194.51k&41.33k&31.81k	& 190.86k \\
		German &520.64k	& -	&217.96k&5.67k&0.21k&	12.20k \\
		Spanish &193.01k&217.46k& -	&159.67k&28.68k	&144.82k \\
		Chinese &40.79k&5.62k&159.73k& -&0.05k&	0.32k\\
		Korean &31.05k&1.37k&28.89k&0.07k& -&56.93k\\
		Dutch &215.18k&25.75k&155.35k&0.66k & 56.92k & -	\\
		\hline
	\end{tabular}
\end{table*}

\subsubsection{Tatoeba}

Tatoeba is a freely available crowd-annotated dataset for language learning. We selected all sentences in English, Spanish, German, Dutch, and Korean. We pruned the dataset to contain only sentences with at least one translation to any of the other languages. The final training set contains 1.36M translation sentence pairs from this source.

\subsubsection{OpenSubtitles}
We augmented our training data by using the 2018 OpenSubtitles dataset. OpenSubtitles is a publicly available dataset based on movie subtitles \cite{opensrt}. We created our training dataset from selected aligned subtitles by taking the unique translations among the first million sentences, for each aligned parallel corpus. We further processed the data by pruning to remove samples with less than 3 words, multiple sentences, or incomplete sentences. The resulting dataset contains 1.9M translation sentence pairs from this source.

\section{Experiments}\label{experiment}
We aim to address the following questions:
\begin{enumerate}
	\item Can syntactic structures be embedded? For multiple languages?
	\item Can parallel corpora be used to learn syntactic structure for low-resource languages?
	\item Does multilingual pre-training improve syntactic embeddings?
\end{enumerate}

We address question 1 in Secs.~\ref{subsec:quality} and~\ref{subsec:dissimilarity} by evaluating the quality of syntactic and semantic embeddings in several ways. Questions 2 and 3  are addressed in Sec.~\ref{subsec:transfer} by studying the transfer-learning performance of syntactic embeddings. 

\subsection{Quality of Syntactic Embeddings}\label{subsec:quality}

We studied the quality of the learned syntactic embeddings by using a nearest-neighbour (NN) method. 

First, we calculated the UPOS sequence of all sentences in the Tatoeba dataset by using a tagger. Sentences were then assigned to distinct groups according to their UPOS sequence, i.e., all sentences belonging to the same group had the same UPOS sequence. 

For all languages except Korean, a held-out test set was created by randomly sampling groups that contained at least 6 sentences. For Korean, all groups containing at least 6 sentences were kept as the test set since the dataset is small. 

During evaluation, we applied max-pooling to the outputs of the encoder to obtain the syntactic embeddings of the held-out sentences\footnote{Evaluation data will be hosted at \url{https://github.com/ccliu2/syn-emb}}.

For each syntactic embedding, we find its top nearest neighbour (1-NN) and top-5 nearest neighbours (5-NN) in the embedding space for the held-out sentences, based on their UPOS group. 

Given $n$ sentences $S = \{s_0, \dots, s_{n-1}\}$ and their embeddings $E = \{e_0, \dots, e_{n-1}\}$, for each $s_i$ there is a set of $k$ gold nearest neighbours $G(i, k) = \{g_0, \dots, g_{k-1}\}$, $G(i, k) \subseteq S$ such that $d(s_i, g) \leq d(s_i, s) \textrm{ for all } g \in G(i, k) \textrm{ and } s \in S \setminus G(i, k)$, where $d(\cdot, \cdot)$ is the cosine distance.

Given embedding $e_i$, we calculate cosine distances $\{d(e_i, e_j) \textrm{ for } e_j \in E, e_j \neq e_i\}$ and sort them into non-decreasing order $d_{j_0} \leq d_{j_1} \leq \dots \leq d_{j_{n-2}}$. We consider the ordering to be unique as the probability of embedding cosine distances being equal is very small. 

The set of embedded $k$-nearest neighbours of $s_i$ is defined as 
\[
N(i, k) = \{s_j \textrm{ for } j \in \{j_0, \dots, j_{k-1}\}\}.
\]
Finally, the $k$-nearest neighbours accuracy for $s_i$ is given by
\begin{equation*}
\frac{|N(i, k) \cap G(i, k)|}{k}.
\end{equation*}

A good embedding model should cluster the embeddings for similar inputs in the embedding space. Hence, the 5-NN test can be seen as an indicator of how cohesive the embedding space is. 

The results are shown in Table \ref{syn_nn}. The differences in the number of groups in each language are due to different availabilities of sentences and sentence-types in the Tatoeba dataset.

The high nearest neighbours accuracy indicates that syntax information was successfully captured by the embeddings. Table~\ref{syn_nn} also shows that the syntactic information of multiple languages was captured by a single embedding model. 


\begin{table}[t]
	\caption{Syntactic Nearest-Neighbour Accuracy (\%) }
	\label{syn_nn}
	\centering
	\begin{tabular}{llll}
		\hline
		{} &  ISO & 1-NN/5-NN &  Total/Groups   \\
		\hline
		English & en & 97.27/93.36 & 2784/160 \\
		German& de & 93.45/86.77 & 1282/91 \\
		Spanish& es & 93.81/86.24 & 1503/81  \\
		Chinese & zh & 71.26/61.44 &  167/22 \\
		Korean& ko & 28.27/18.40 & 527/40\\
		Dutch& nl & 74.17/51.71 & 3171/452 \\
		\hline
	\end{tabular}
\end{table}

\subsubsection{Language Model}
A number of recent works \cite{hewitt2019, yoav2019} have probed language models to determine if they contain syntactic information. We applied the same nearest neighbours experiment (with the same test sets) on a number of existing language models: Universal Sentence Encoder (USE) \cite{USE}, LASER, and BERT. For USE we used models available from TensorHub\footnote{https://www.tensorflow.org/hub}. For LASER we used models and created embeddings from the official repository \footnote{https://github.com/facebookresearch/LASER}. 

For BERT, we report the results using max (BERT$_{max}$) and average-pooling (BERT$_{avg}$), obtained from the BERT embedding toolkit\footnote{https://github.com/imgarylai/bert-embedding} with the multilingual cased model (104 languages, 12-layers, 768-hidden units, 12-heads), and `pooled-output' (BERT$_{output}$) from the TensorHub version of the model with the same parameters.

We computed the nearest neighbours experiment for all languages in the training data for the above models. The results are shown in Table~\ref{syn_nn_appxb}. The results show that general purpose language models do capture syntax information, which varies greatly across languages and models. 

\begin{table*}
	\caption{Syntactic Nearest-Neighbour for Language Models (\%)}
	\label{syn_nn_appxb}
	\centering
	\begin{tabular}{lllllll}
		\hline
		{} & English & German & Spanish & Chinese & Korean & Dutch  \\
		Model & 1-NN/5-NN & 1-NN/5-NN & 1-NN/5-NN & 1-NN/5-NN & 1-NN/5-NN & 1-NN/5-NN   \\
		\hline
		USE & 71.83/55.68 & 59.87/44.26 &  53.05/38.06  & 39.23/30.18 & 21.22/12.43 & 28.66/12.77\\
		BERT$_{max}$ & \textbf{90.19/86.36} & \textbf{83.66/77.63} & \textbf{83.89/79.92} & \textbf{67.96/68.40} &20.30/11.92& 37.67/19.51\\
		BERT$_{avg}$ & 89.06/84.70 & 79.54/74.82 & 78.24/75.61 & 65.75/67.07 & 20.30/11.47 & 37.04/19.46\\
		BERT$_{output}$ & 77.75/63.44 & 66.20/51.89 & 65.21/50.41 & 52.49/46.34 &16.39/10.98 & 24.27/10.67 \\
		LASER & 86.33/76.66 & 76.56/62.88 & 72.49/59.72 & 56.89/45.15 & \textbf{26.63/15.90} & \textbf{50.75/31.00}\\
		\hline
	\end{tabular}
\end{table*}

The nearest neighbours accuracy of our syntactic embeddings in Table~\ref{syn_nn} significantly outperforms the general purpose language models. Arguably these language models were trained using different training data. However, this is a reasonable comparison because many real-world applications rely on released pre-trained language models for syntactically related information. Hence, we want to show that we can use much smaller models trained with direct supervision, to obtain syntactic embeddings with similar or better quality. Nonetheless, the training method used in this work can certainly be extended to architectures similar to BERT or USE.


\subsection{Functional Dissimilarity}
\label{subsec:dissimilarity}
The experiments in the previous section showed that the proposed syntactic embeddings formed cohesive clusters in the embedding space, based on UPOS sequence similarities. We further studied the spatial relationships within the embeddings.

\textit{Word2Vec} \cite{word2vec} examined spatial relationships between embeddings and compared them to the semantic relationships between words. Operations on vectors in the embedding space such as $King - Man + Woman = Queen$ created vectors that also correlated with similar operations in semantics. Such semantic comparisons do not directly translate to syntactic embeddings. However, syntax information shifts with edits on POS sequences. Hence, we examined the spatial relationships between syntactic embeddings by comparing their cosine similarities with the edit distances between UPOS sequence pairs. 

Given $n$ UPOS sequences $U = \{u_0,...,u_{n-1}\}$, we compute the matrix $L \in \mathbb{R}^{n \times n}$, where $l_{ij} = l(u_i, u_j)$, the complement of the normalized Levenshtein distance between $u_i$ and $u_j$.

Given the set of embedding vectors $\{e_0,...,e_{n-1}\}$ where $e_i$ is the embedding for sentence $s_i$, we also compute $D \in \mathbb{R}^{n \times n}$, where $d_{ij} = d(e_i, e_j)$. We further normalize $d_{ij}$ to be within $[0, 1]$ by min-max normalization to obtain $\hat{D} = \minMax(D)$.

Following \cite{pip}, we define the \emph{functional dissimilarity score} by
\begin{equation*}
\frac{\|L - \hat{D}\|_{\rm{F}}}{n}.
\end{equation*}

Intuitively, UPOS sequences that are similar (smaller edit distance) should be embedded close to each other in the embedding space, and embeddings that are further away should have dissimilar UPOS sequences. Hence, the functional dissimilarity score is low if the relative changes in UPOS sequences are reflected in the embedding space. The score is high if such changes are not reflected. 

The functional dissimilarity score was computed using sentences from the test set in CoNLL 2017 Universal Dependencies task \cite{conll} for the relevant languages with the provided UPOS sequences. Furthermore, none of the evaluated models, including the proposed method, were trained with CoNLL2017 data. 

We compared the functional dissimilarity scores of our syntactic representations against embeddings obtained from BERT and LASER, to further demonstrate that simple network structures with explicit supervision may be sufficient to capture syntactic structure. All the results are shown in Table~\ref{pip}. We only show the best (lowest) results from BERT.

\begin{table*}
	\caption{Functional Dissimilarity Scores (Lower is Better)}
    \label{pip}
    \centering
    \begin{tabular}{lllllll}
        \hline
        Model & English & German & Spanish & Chinese & Korean & Dutch  \\
        \hline
    	BERT$_{avg}$ & 0.3463 & 0.3131 & 0.2955 & 0.2935 & 0.3001 & 0.3131\\
        LASER & 0.1602 & 0.1654 & 0.2074 & 0.3099 & 0.2829 & 0.1654\\
        Proposed Work & 0.1527 & 0.1588 & 0.1588 & 0.2267 & 0.2533 & 0.1588 \\
        \hline
    \end{tabular}
\end{table*}

\subsection{Transfer Performance of Syntactic Embeddings}\label{subsec:transfer}

Many NLP tasks utilize POS as features, but human annotated POS sequences are difficult and expensive to obtain. Thus, it is important to know if we can learn sentences-level syntactic embeddings for low-sources languages without treebanks.

We performed zero-shot transfer of the syntactic embeddings for French, Portuguese and Indonesian. French and Portuguese are simulated low-resource languages, while Indonesian is a true low-resource language. We reported the 1-NN and 5-NN accuracies for all languages using the same evaluation setting as described in the previous section. The results are shown in Table~\ref{transfer_nn} (top). 

We also fine-tuned the learned syntactic embeddings on the low-resource language for a varying number of training data and languages. The results are shown in Table \ref{transfer_nn} (bottom). In this table, the low-resource language is denoted as the `source', while the high-resource language(s) is denoted as the `target'.  With this training method, no UPOS tag information was provided to the model for the `source' languages, where supervising information comes solely from parallel sentences and UPOS tags in high-resource languages.

The results show that for a new language (French and Portuguese) that is similar to the family of pre-training languages, there are two ways to achieve higher 1-NN accuracy. If the number of unique sentences in the new language is small, accuracy can be improved by increasing the size of the parallel corpora used to fine-tune. If only one parallel corpus is available, accuracy can be improved by increasing the number of unique sentence-pairs used to fine-tune.

For a new language that is dissimilar to the family of pre-training languages, e.g. Indonesian in Table~\ref{transfer_nn}, the above methods only improved nearest neighbours accuracy slightly. This may be caused by differing data distribution or by tagger inaccuracies. The results for Indonesian do indicate that some syntactic structure can be learned by using our method, even for a dissimilar language.

A future direction is to conduct a rigorous analysis of transfer learning between languages from the same versus different language families.

\begin{table}
	\caption{Syntactic Nearest-Neighbour on New languages (\%)}
	\label{transfer_nn}
	\centering
	\begin{tabular}{lll}
		\hline
		Lang (ISO)& {1-NN/5-NN} & Total/Group \\
		\hline
		French (fr)  & 35.86/22.11 & 6816/435 \\
		Protuguese (pt) & 48.29/23.15 & 4608/922\\
		Indonesian (id)  & 21.00/35.92 & 657/59\\
		\hline
		\hline
		\multicolumn{3}{c}{}{Number of Parallel Sentence Pairs}  \\
		Source -Target(s)   & 2k & 10k  \\
		ISO  & 1-NN/5-NN & 1-NN/5-NN   \\
		\hline
		fr-en  &  47.37/32.18 & 58.41/42.87  \\
		fr-(en,es)  & 46.82/31.92 & 58.01/42.65 \\
		\hline
		pt-en  & 56.75/30.14 & 64.52/36.94  \\
		pt-(en,es)  & 57.94/30.63 & 65.00/37.06 \\
		\hline
		id-en  &  27.09/47.64  & 31.35/56.01\\
		\hline
	\end{tabular}
\end{table}

%
%

\section{Conclusion}
We examined the possibility of creating syntactic embeddings by using a multilingual method based on sequence-to-sequence models. In contrast to prior work, our method only requires parallel corpora and UPOS tags in the target language.

We studied the quality of learned embeddings by examining nearest neighbours in the embedding space and investigating their functional dissimilarity. These results were compared against recent state-of-the-art language models. We also showed that pre-training with a parallel corpus allowed the syntactic embeddings to be transferred to low-resource languages via few-shot fine-tuning. 

Our evaluations indicated that syntactic structure can be learnt by using simple network architectures and explicit supervision. Future directions include improving the transfer performance for low-resource languages, disentangling semantic and syntactic embeddings, and analyzing the effect of transfer learning between languages belong to the same versus different language families. 

\bibliography{emnlp-ijcnlp-2019}

\begin{thebibliography}{23}
\expandafter\ifx\csname natexlab\endcsname\relax\def\natexlab#1{#1}\fi

\bibitem[{Artetxe and Schwenk(2018)}]{artetxe2018}
Mikel Artetxe and Holger Schwenk. 2018.
\newblock Massively multilingual sentence embeddings for zero-shot
  cross-lingual transfer and beyond.
\newblock \emph{arXiv preprint arXiv:1812.10464}.

\bibitem[{Cer et~al.(2018)Cer, Yang, yi~Kong, Hua, Limtiaco, John, Constant,
  Guajardo-Cespedes, Yuan, Tar, Sung, Strope, and Kurzweil}]{USE}
Daniel Cer, Yinfei Yang, Sheng yi~Kong, Nan Hua, Nicole Limtiaco, Rhomni~St.
  John, Noah Constant, Mario Guajardo-Cespedes, Steve Yuan, Chris Tar,
  Yun-Hsuan Sung, Brian Strope, and Ray Kurzweil. 2018.
\newblock Universal sentence encoder.
\newblock \emph{CoRR}, abs/1803.11175.

\bibitem[{Chen et~al.(2019)Chen, Tang, Wiseman, and Gimpel}]{chen2019}
Mingda Chen, Qingming Tang, Sam Wiseman, and Kevin Gimpel. 2019.
\newblock A multi-task approach for disentangling syntax and semantics in
  sentence representations.
\newblock In \emph{NAACL2019}.

\bibitem[{Conneau and Kiela(2018)}]{SentEval}
Alexis Conneau and Douwe Kiela. 2018.
\newblock Senteval: An evaluation toolkit for universal sentence
  representations.
\newblock \emph{CoRR}, abs/1803.05449.

\bibitem[{Conneau et~al.(2017)Conneau, Lample, Ranzato, Denoyer, and
  J{\'e}gou}]{conneau2017word}
Alexis Conneau, Guillaume Lample, Marc'Aurelio Ranzato, Ludovic Denoyer, and
  Herv{\'e} J{\'e}gou. 2017.
\newblock Word translation without parallel data.
\newblock \emph{arXiv preprint arXiv:1710.04087}.

\bibitem[{Devlin et~al.(2018)Devlin, Chang, Lee, and Toutanova}]{BERT}
Jacob Devlin, Ming-Wei Chang, Kenton Lee, and Kristina Toutanova. 2018.
\newblock Bert: Pre-training of deep bidirectional transformers for language
  understanding.
\newblock \emph{arXiv preprint arXiv:1810.04805}.

\bibitem[{Fang and Cohn(2016)}]{meng2016}
Meng Fang and Trevor Cohn. 2016.
\newblock Learning when to trust distant supervision: An application to
  low-resource {POS} tagging using cross-lingual projection.
\newblock In \emph{Proceedings of the 20th SIGNLL Conference on Computational
  Natural Language Learning}.

\bibitem[{Goldberg(2019)}]{yoav2019}
Yoav Goldberg. 2019.
\newblock Assessing bert's syntactic abilities.
\newblock \emph{arXiv preprint arXiv:1901.05287}.

\bibitem[{Hewitt and Manning(2019)}]{hewitt2019}
John Hewitt and Christopher~D. Manning. 2019.
\newblock A structural probe for finding syntax in word representations.
\newblock In \emph{NAACL2019}.

\bibitem[{Kingma and Ba(2015)}]{ADAM}
Diederik~P. Kingma and Jimmy Ba. 2015.
\newblock Adam: A method for stochastic optimization.
\newblock \emph{CoRR}, abs/1412.6980.

\bibitem[{Lample and Conneau(2019)}]{XlangPretrain}
Guillaume Lample and Alexis Conneau. 2019.
\newblock Cross-lingual language model pretraining.
\newblock \emph{CoRR}, abs/1901.07291.

\bibitem[{Linzen et~al.(2016)Linzen, Dupoux, and Goldberg}]{linzen2016}
Tal Linzen, Emmanuel Dupoux, and Yoav Goldberg. 2016.
\newblock Assessing the ability of lstms to learn syntax-sensitive
  dependencies.
\newblock \emph{Transactions of the Association for Computational Linguistics}.

\bibitem[{Lison and Tiedemann(2016)}]{opensrt}
Pierre Lison and J{\"o}rg Tiedemann. 2016.
\newblock Opensubtitles2016: Extracting large parallel corpora from movie and
  tv subtitles.

\bibitem[{Marvin and Linzen(2018)}]{marvin2018}
Rebecca Marvin and Tal Linzen. 2018.
\newblock Targeted syntactic evaluation of language models.
\newblock In \emph{EMNLP2018}.

\bibitem[{Nivre et~al.(2017)Nivre, Agi{\'c}, Ahrenberg et~al.}]{conll}
Joakim Nivre, {\v{Z}}eljko Agi{\'c}, Lars Ahrenberg, et~al. 2017.
\newblock Universal dependencies 2.0. lindat/clarin digital library at the
  institute of formal and applied linguistics, charles university, prague.

\bibitem[{Plank and Agic(2018)}]{plank2018}
Barbara Plank and Zeljko Agic. 2018.
\newblock Distant supervision from disparate sources for low-resource
  part-of-speech tagging.
\newblock In \emph{EMNLP2018}.

\bibitem[{Qi et~al.(2018)Qi, Dozat, Zhang, and Manning}]{qi2018universal}
Peng Qi, Timothy Dozat, Yuhao Zhang, and Christopher~D. Manning. 2018.
\newblock \href {https://nlp.stanford.edu/pubs/qi2018universal.pdf} {Universal
  dependency parsing from scratch}.
\newblock In \emph{Proceedings of the {CoNLL} 2018 Shared Task: Multilingual
  Parsing from Raw Text to Universal Dependencies}, pages 160--170, Brussels,
  Belgium. Association for Computational Linguistics.

\bibitem[{Schwenk(2018)}]{minning}
Holger Schwenk. 2018.
\newblock Filtering and mining parallel data in a joint multilingual space.
\newblock In \emph{ACL}.

\bibitem[{Sennrich et~al.(2016)Sennrich, Haddow, and
  Birch}]{sennrich-etal-2016-neural}
Rico Sennrich, Barry Haddow, and Alexandra Birch. 2016.
\newblock \href {https://doi.org/10.18653/v1/P16-1162} {Neural machine
  translation of rare words with subword units}.
\newblock In \emph{Proceedings of the 54th Annual Meeting of the Association
  for Computational Linguistics (Volume 1: Long Papers)}, pages 1715--1725,
  Berlin, Germany. Association for Computational Linguistics.

\bibitem[{Sutskever et~al.(2014)Sutskever, Vinyals, and Le}]{seq2seq}
Ilya Sutskever, Oriol Vinyals, and Quoc~V. Le. 2014.
\newblock Sequence to sequence learning with neural networks.
\newblock In \emph{NIPS}.

\bibitem[{Wang et~al.(2019)Wang, Singh, Michael, Hill, Levy, and Bowman}]{GLUE}
Alex Wang, Amanpreet Singh, Julian Michael, Felix Hill, Omer Levy, and
  Samuel~R. Bowman. 2019.
\newblock {GLUE}: A multi-task benchmark and analysis platform for natural
  language understanding.
\newblock In the Proceedings of ICLR.

\bibitem[{Warstadt et~al.(2018)Warstadt, Singh, and Bowman}]{cola}
Alex Warstadt, Amanpreet Singh, and Samuel~R Bowman. 2018.
\newblock Neural network acceptability judgments.
\newblock \emph{arXiv preprint arXiv:1805.12471}.

\bibitem[{Yin and Shen(2018)}]{pip}
Zi~Yin and Yuanyuan Shen. 2018.
\newblock On the dimensionality of word embedding.
\newblock In \emph{Advances in Neural Information Processing Systems}, pages
  887--898.

\end{thebibliography}
\bibliographystyle{acl_natbib}

\newpage

\end{document}